\documentclass[manuscript,nonacm]{acmart}

\usepackage{booktabs}
\usepackage{comment}

\usepackage{graphicx}
\usepackage{multirow}
\usepackage{adjustbox}
\usepackage{caption,subcaption,graphicx}

\AtBeginDocument{%
  }

\begin{document}

\title[Exploring a Multimodal Fusion-based Deep Learning Network for Detecting Facial Palsy]{Exploring a Multimodal Fusion-based Deep Learning Network for Detecting Facial Palsy}


\author{Heng Yim Nicole Oo}
\email{nicoleoo.2021@scis.smu.edu.sg}
\affiliation{%
  \institution{Singapore Management University}
  \city{Singapore}
  \country{Singapore}
}

\author{Min Hun Lee}
\email{mhlee@smu.edu.sg}
\affiliation{%
  \institution{Singapore Management University}
  \city{Singapore}
  \country{Singapore}
}

\author{Jeong Hoon Lim}
\affiliation{%
  \institution{National University of Singapore}
  \city{Singapore}
  \country{Singapore}}

\renewcommand{\shortauthors}{Oo et al.}

\begin{abstract}
Algorithmic detection of facial palsy offers the potential to improve current practices, which usually involve labor-intensive and subjective assessment by clinicians. In this paper, we present a multimodal fusion-based deep learning model that utilizes unstructured data (i.e. an image frame with facial line segments) and structured data (i.e. features of facial expressions) to detect facial palsy. We then contribute to a study to analyze the effect of different data modalities and the benefits of a multimodal fusion-based approach using videos of 21 facial palsy patients. Our experimental results show that among various data modalities (i.e. unstructured data - RGB images and images of facial line segments and structured data - coordinates of facial landmarks and features of facial expressions), the feed-forward neural network using features of facial expression achieved the highest precision of 76.22 while the ResNet-based model using images of facial line segments achieved the highest recall of 83.47. When we leveraged both images of facial line segments and features of facial expressions, our multimodal fusion-based deep learning model slightly improved the precision score to 77.05 at the expense of a decrease in the recall score. 
\end{abstract}



\maketitle

\section{Introduction \& Related Work}
Facial palsy patients usually suffer from symptoms, such as drooping mouth and eyebrows, difficulty in closing an eyelid, and drooling. Facial palsy has serious consequences on patients, such as diminished feeding function, psychological distress, and social withdrawal \cite{kosins2007}. For diagnosis of facial palsy, clinicians usually conduct observation-based physical examinations \cite{heckmann2019diagnosis}. 
However, it is challenging to quantify symptom intensity and variation, measure changes in these symptoms between visits for a single patient, and measure differences in symptoms across different patients at the same time \cite{hsu2018deep}. 

To address this challenge, researchers have explored various algorithmic approaches to detect facial palsy \cite{hsu2018deep,Ngo2016QuantitativeAO,kim2015smartphone,Wang2014AutomaticRO}. These approaches broadly fall into two categories: 1) those employing machine learning models with manual feature extraction and 2) those that leverage deep learning-based models.

For approaches with manual features, Ngo et al. \cite{Ngo2016QuantitativeAO} proposed a frequency-based technique using limited-orientation modified circular Gabor filters (LO-MCGFS) to magnify desired frequencies in dataset images and extract features from rotation invariant texture regions for classifying facial palsy. In addition, researchers explored to train a data-driven model \cite{kim2015smartphone,Wang2014AutomaticRO,He2009} to detect facial key points and computed features, such as the displacement ratio between left and right halves of the face or the motion information of facial regions.

Alternatively, researchers discussed the limitations of using manual features and leveraged RGB images or images with facial line segments to train deep learning-based models (e.g. 3-dimensional or hierarchy convolutional neural network) for detecting facial palsy \cite{hsu2018deep,liu2020region,storey20193dpalsynet}. However, it remains unclear whether deep learning-based models using unstructured data can outperform the approaches using structured data of manual features and if combining unstructured data and structured data of manual features can improve the performance of detecting facial palsy. 

In this work, we contribute to a study that analyzes the performance of using diverse data modalities (i.e. unstructured data: RGB images and images with facial line segments and structured data: coordinates of facial landmarks and features of facial expression). In addition, we present a multimodal fusion-based approach (Figure \ref{fig:model_architecture}) to explore the benefits of integrating different data modalities for detecting facial palsy. According to the evaluation study using the dataset of 21 facial palsy patients \cite{hsu2018deep}, among four data modalities, we found that the feed-forward neural network model using features of facial expression achieved the highest average precision of 76.22 while 
the ResNet-based model using images with facial line segments achieved the highest average recall of 83.47. When we leveraged multimodal fusion-based deep learning models using both features of facial expression and images with facial line segments, the model achieved higher performance than using only unstructured data of RGB images and facial line segments \cite{hsu2018deep}. In addition, our multimodal fusion-based models slightly improved an average precision score by 0.78\% at the expense of reducing an average recall by 5-9\%. We discuss the potential and limitations of a multimodal fusion-based approach for detecting facial palsy.

\section{Methods} \label{section:methods}
In this section, we describe how we process RGB images to obtain four independent data modalities and discuss various model architectures to detect facial palsy.

\subsection{Data Processing \& Modalities}\label{sect:data_modal}
\begin{figure}[htp!]
\centering
\begin{subfigure}[t]{0.2\columnwidth}
\centering
  \includegraphics[width=1.0\columnwidth]{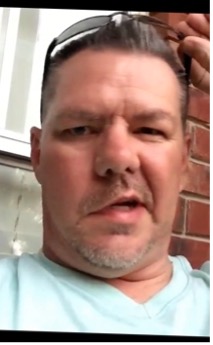}
  \caption{}
  \label{fig:data_rgb}
\end{subfigure}
\begin{subfigure}[t]{0.19\columnwidth}
\centering
  \includegraphics[width=1.0\columnwidth]{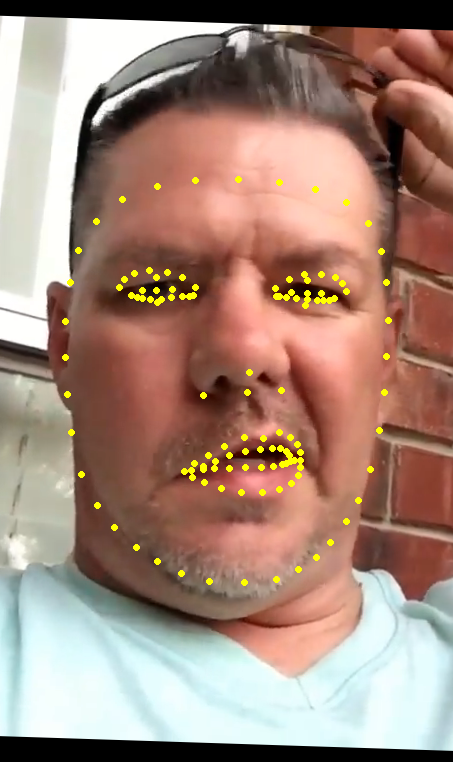}
  \caption{}
  \label{fig:data_3d}
\end{subfigure}
\begin{subfigure}[t]{0.29\columnwidth}
\centering
  \includegraphics[width=1.0\columnwidth]{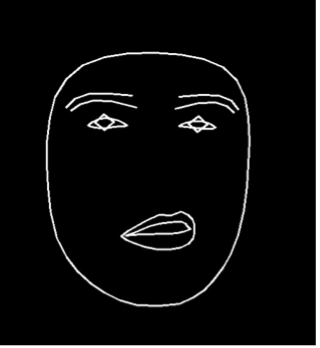}
  \caption{}
  \label{fig:data_bnw}
\end{subfigure}
\caption{(a) A sample RGB image of a patient with facial palsy. (b) 125 3-dimensional coordinates of eyes, nose, and mouth regions overlaid on the RGB image, and (c) line segments of facial expressions on the BnW image}\label{fig:sample_data}
\end{figure}

To develop an AI model, we processed raw RGB image frames of videos using the facial landmark estimation model \cite{lugaresi2019mediapipe} to obtain four data modalities: 1) raw RGB images (Figure \ref{fig:data_rgb}), 2) facial landmark coordinates (Figure \ref{fig:data_3d}), 3) features of facial expressions, 4) black and white (BnW) images with line segments representing the facial silhouette and local regions (Figure \ref{fig:data_bnw}).

For facial landmark coordinates, we include only 125 2-dimensional coordinates of eyes, nose, and mouth regions among 478 3-dimensional coordinates of facial landmarks from the model \cite{lugaresi2019mediapipe}. The resulting facial landmark coordinates are a matrix of $\mathbf{m} \in \mathbb{R}^{125x2}$.

Feature vectors of facial expressions contain 52 distinct attributes, each representing the extent of a facial expression (e.g. the extent to which each eye is closed, or to which the mouth is opened) in a float value [0, 1]. The resulting vector of features of facial expressions is $\mathbf{b} \in \mathbb{R}^{52}$.

To obtain the BnW line segment images, we first used the facial landmark estimation model to generate contours of the detected face, eyebrows, and eyes, then plotted them in white against a solid black background (Figure \ref{fig:data_bnw}).

\begin{figure*}[htp]
\centering
\includegraphics[width=0.8\linewidth]{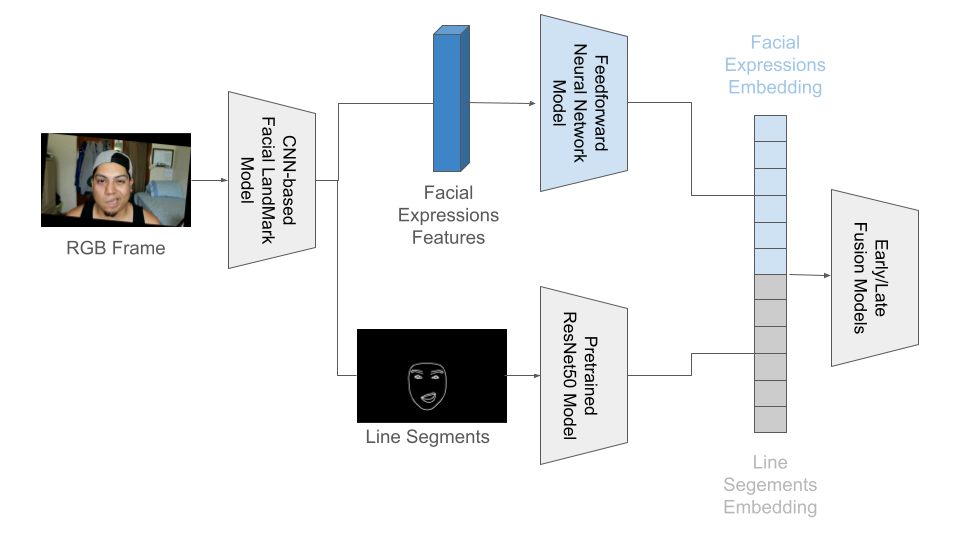}
\caption{Our early fusion model integrates facial expression-based embedding from a feedforward neural network with line segment-based embedding from a pre-trained ResNet50 model to detect a patient with facial palsy}\label{fig:model_architecture}
\end{figure*}

\subsection{Model Architectures}
Given four data modalities, we explored different model architectures (i.e. feed-forward neural networks, ResNet50-based models) tailored to each data modality and combined modalities.

\subsubsection{Problem Formulation}
We formulated the detection of facial palsy as a binary classification task, where the presence of facial palsy is defined as either strong palsy intensity in any one region (Eyes or Mouth) or slight intensity in both regions (slight palsy intensity in both Eyes and Mouth). This approach allows us to indicate either the presence or absence of facial palsy for each image. 

Given the binary classification task, we applied the Binary Cross-Entropy (BCE) as the objective loss function.

\begin{displaymath}
BCE = -{(y\log(p) + (1 - y)\log(1 - p))}
\end{displaymath}

\subsubsection{Feed-forward Neural Network Model}
Feed-forward Neural Network (FNN) models were trained to perform classification using facial landmark coordinates and feature vectors of facial expression separately. Both FNN models consisted of four fully connected layers. We applied a ReLU activation function on the first three fully connected layers and a sigmoid function on the last layer to output class probabilities. In the first layer, we applied a dropout layer with a probability of 0.5. In the second layer, we applied a 1-dimensional batch norm to make the model more stable. We utilized the hidden unit sizes of 128, 64, 32, and 2 for the facial landmark coordinates and those of 64, 32, 10, and 2 for the feature vectors of facial expression. 

\subsubsection{ResNet50-based model}
For the ResNet50-based model, we fine-tuned the pre-trained ResNet50 model \cite{he2016deep} to perform classification using raw RGB images or BnW images with line segments. The final layer in the pre-trained ResNet50 model was modified to include a fully-connected layer of 512 hidden units followed by a ReLU activation function, a dropout layer, and a 1-dimensional batch norm layer and another fully-connected layer with a sigmoid activation at the end to output class probabilities. 

\subsubsection{Early \& Late Fusion Models}
With the goal of improving facial palsy detection, we explored a multimodal fusion approach  \cite{baltruvsaitis2018multimodal} using unstructured and unstructured data. Specifically, we utilized model-agnostic approaches of 1) early fusion that integrates two data modalities after extracting their features and concatenating their features and 2) late fusion that integrates the outputs of the models using each of the modalities (Figure 
\ref{fig:model_architecture}).

For our fusion models, we utilized the models that were independently trained using a single modality (Section \ref{sect:data_modal}) and then empirically identified two modalities that achieved the highest performance in detecting facial palsy. 

For the early fusion, we extracted the feature vectors from two models that were independently trained using a single modality and concatenated these feature vectors to train an early fusion model (Figure \ref{fig:model_architecture}). We empirically chose the intermediate layer of a model using a single model to extract a feature vector. The early fusion model consists of four fully connected layers with ReLU activation functions. Except for the third fully connected layer, we applied dropout layers with a probability of 0.5. In the third fully connected layer, we applied a batch norm layer. 

For the late fusion, we computed the average of the output probabilities from two models that were independently trained using a single modality (i.e. the features of facial expressions or BnW images with line segments) and generated the model predictions on the class with the highest mean probability.

\section{Experiments} \label{section:experiments}
\subsection{Dataset}
Our experiments were conducted using the YouTube Facial Palsy (YFP) dataset by \cite{hsu2018deep}, which has been labeled by three independent clinicians. It consists of 31 videos collected from 21 facial palsy patients, and some patients have provided multiple videos. In each video, the patient talks to the camera, recording variations in their facial expressions over time. Each video has been converted to an image sequence with a sampling rate of 6FPS. For each image in the image sequence, the dataset contains labels indicating the intensity of palsy in two facial regions (i.e. the eyes and mouth) where it can be observed.

\subsection{Evaluation}
For the evaluation of different data modalities and model architectures, we applied the leave-one-patient-out (LOPO) cross-validation, in which we trained the model using data except for held-out data from a patient for testing and we repeated this process over 21 patients. For each patient, we recorded the F1-score, precision, and recall and then averaged these metrics for each data modality and model architecture. 

To understand the benefits of our early and late fusion-based deep learning models using unstructured data and structured data, we also implemented a ResNet-based model that utilized two formats of unstructured data (i.e. RGB images and BnW images with line segments) followed by \cite{hsu2018deep}. 

For training a model, we applied the stochastic gradient descent (SGD) algorithm to optimize model parameters and explored different learning rates. All models were trained for 15 epochs except for the ResNet-based model using two unstructured data formats, which was trained for 8 epochs. We empirically found that the optimal learning rate of the FFN models was 0.01 and that of the ResNet-based models was 0.001.

\section{Results \& Discussion} \label{section:results}
Table \ref{tab:results} summarises our experimental results over different data modalities and model architectures.

\begin{table*}[htp]
\centering
\caption{Experimental results of different data modalities and model architectures}
\label{tab:results}
\resizebox{\textwidth}{!}{%
\begin{tabular}{cccccc} \toprule
\textbf{Data Modality} &
  \textbf{Model} &
  \textbf{Average F1} &
  \textbf{Average Precision} &
  \textbf{Average Recall} \\ \midrule
Coordinates  & Feed-forward Neural Network      & 72.23          & 74.40          & 83.28      \\
Features of Facial Expressions  & Feed-forward Neural Network      & 71.21          & {76.22}          & 79.00     \\
RGB Images   & ResNet50-based Model & 64.23          & {76.14} & 62.04      \\ 
BnW LineSegment Images       & ResNet50-based Model & \textbf{72.85} & 75.57          & \textbf{83.47}  \\ \midrule \midrule
\begin{tabular}[c]{@{}c@{}}BnW LineSegment Images +\\ RGB Images\end{tabular}  & ResNet50-based Model & 65.95               & 76.95               & 67.42                \\ \midrule
\begin{tabular}[c]{@{}c@{}}Features of Facial Expressions +\\BnW LineSegment Images \end{tabular} &
  \begin{tabular}[c]{@{}c@{}}Early Fusion Model\\ (ResNet50 + Feedforward Model)\end{tabular} &
  69.62 &
  \textbf{77.05} &
  74.71 &
   \\ \midrule
\begin{tabular}[c]{@{}c@{}}Features of Facial Expressions +\\BnW LineSegment Images\end{tabular} &
  \begin{tabular}[c]{@{}c@{}}Late Fusion Model\\(ResNet50 + Feedforward Model)\end{tabular} &
  71.76 &
  {77.00} &
  78.66 \\ \bottomrule 
\end{tabular} 
} 
\end{table*}

When we used single data modalities, the feed-forward neural network (FNN) model using facial landmark coordinates achieved an average F1-score of 72.23\%, average precision of 74.40\%, and average recall of 83.28\%. The FNN model using features of facial expressions achieved an average F1-score of 71.21\%, average precision of 76.22\%, and average recall of 79.00\%. The ResNet-based model using RGB images achieved an average F1-score of 64.23\%, average precision of 76.14\%, and average recall of 62.04\%.  The ResNet-based model using BnW Line Segment images achieved an average F1-score of 72.85\%, average precision of 75.57\%, and average recall of 83.47\%. 

Among structured data (i.e. coordinates of facial landmarks and features of facial expressions), the FNN model using coordinates of facial landmarks achieved a higher average F1-score, a higher average recall, and a lower average precision than the FNN model using features of facial expressions. Among unstructured data (i.e. RGB images and BnW images with line segments, the ResNet50-based model using BnW images with line segments achieved a higher average F1-score, a slightly lower precision, and a higher average recall than that using RGB images. 

Among all data modalities, \textbf{the ResNet-based model using BnW images with line segments achieved the highest F1-score of 72.85 and the highest recall of 83.47} and \textbf{the FNN model using features of facial expression achieved the highest precision of 76.14}. Thus, we further explored the benefit of integrating the data modalities of BnW images with line segments and the features of facial expressions.

For multiple data modalities, the ResNet-based model using BnW images with line segments and RGB images achieved an average F1-score of 65.95\%, average precision of 76.95\%, and average recall of 67.42\%. The early fusion model that utilized the ResNet-based model for BnW images with line segments and the FNN model for the features of facial expressions achieved an average F1-score of 69.62\%, average precision of 77.05\%, and average recall of 74.71\%.
The late fusion model that utilized the ResNet-based model for BnW images with line segments and the FNN model for the features of facial expressions achieved an average F1-score of 71.76\%, average precision of 77.00\%, and average recall of 78.66\%.

When we utilize structured data (i.e. features of facial expression) and unstructured data (i.e. BnW images with line segments), we found that \textbf{both early and late fusion-based models outperformed the ResNet50-based model using only unstructured data of RGB images and BnW images with line segments by around 3-5\% of average F1-score}. Compared to the ResNet-based model using only RGB images with 64.23\% average F1-score, our early and late fusion models achieved 5\% higher average F1-scores around 70\% average F1-score. Compared to the ResNet-based model using only BnW images with line segments, \textbf{both early and late fusion models slightly improved an average precision by 0.78\% average precision at the expense of reducing an average F1-score by 1-3\% F1-score and an average recall by 5-9\%}. 

Overall, our experimental results showed that the benefit of processing RGB images to generate unstructured data (i.e. BnW images with line segments) or structured data (e.g. features of facial expression) and combining both unstructured and structured data instead of just relying on RGB images or combining only unstructured data. However, as we only observed marginal improvement of average precision score in our early and late fusion models compared to the ResNet-based model using BnW images with line segments, we plan to further explore different fusion model architectures and attention-based approach \cite{han2022survey}. In addition, it is important to study how to make an AI output explainable \cite{doshi2017towards,lee2023exploring} to a clinician and support effective assistance to improve the practices of a clinician \cite{lee2021human,ghassemi2021false}.

\section{Conclusion} 
In this work, we contributed to a study that provided a comprehensive analysis of the performances of different AI models to detect facial palsy using different data modalities (i.e. unstructured data: RGB images and BnW images with facial line segments and structured data: coordinates of facial landmarks and features of facial expressions). 

According to the experiment using the public dataset of 21 facial palsy patients, our study demonstrated that the models using either BnW images with facial line segments or features of facial expressions outperformed the model using raw RGB images. Thus, it is beneficial to utilize processed data (i.e. BnW images with facial line segments or features of facial expressions) instead of using raw RGB image data. In addition, our early and late fusion models using features of facial expressions and BnW images with line segments outperformed the model using only unstructured data of RGB images and BnW images with line segments. Our study discusses the potential of combining diverse data modalities to improve the performance of an AI model to detect facial palsy.

\bibliographystyle{ACM-Reference-Format}
\bibliography{main}

\appendix

\end{document}